\documentclass[letterpaper, 10 pt, conference]{ieeeconf}  

\IEEEoverridecommandlockouts                              

\RequirePackage{luatex85}
\usepackage{times}

\usepackage{soul}
\usepackage{url}
\usepackage[draft]{hyperref}
\usepackage[utf8]{inputenc}
\usepackage[T1]{fontenc}
\usepackage[small]{caption}
\usepackage{amsmath}
\usepackage{booktabs}
\usepackage{pgf}

\usepackage[style=ieee,	sorting=none,bibencoding=utf8, maxcitenames=1, mincitenames=1, minbibnames=1, maxbibnames=1, backend=biber, citestyle=numeric-comp, isbn=false,doi=false, url=false, natbib=true]{biblatex}

\AtEveryBibitem{%
	\ifentrytype{book}{
		\clearfield{url}%
		\clearfield{urldate}%
		\clearfield{review}
		\clearfield{series}
		\clearfield{note}
		\clearlist{location}
		\clearname{editor}
		\clearfield{pages}
	}{}
	\ifentrytype{collection}{
		\clearfield{url}%
		\clearfield{urldate}%
		\clearfield{review}%
		\clearfield{series}
		\clearfield{note}
		\clearlist{location}
		\clearname{editor}
		\clearfield{pages}
	}{}
	\ifentrytype{incollection}{
		\clearfield{url}%
		\clearfield{urldate}%
		\clearfield{review}%
		\clearfield{series}
		\clearfield{note}
		\clearlist{location}
		\clearfield{pages}
		\clearname{editor}
	}{}
	\ifentrytype{inproceedings}{
		\clearlist{location}
		\clearfield{url}%
		\clearfield{urldate}%
		\clearfield{review}%
		\clearfield{series}
		\clearfield{publisher}
		\clearfield{note}	
		\clearfield{editor}
		\clearfield{pages}
		\clearname{editor}
	}{}
	\ifentrytype{article}{
		\clearlist{location}
		\clearfield{url}%
		\clearfield{urldate}%
		\clearfield{review}%
		\clearfield{series}
		\clearfield{publisher}
		\clearfield{note}	
		\clearfield{pages}
		\clearname{editor}
	}{}
}

\makeatletter
\newcommand\fs@spaceruled{\def\@fs@cfont{\bfseries}\let\@fs@capt\floatc@ruled
	\def\@fs@pre{\vspace{1\baselineskip}\hrule height.8pt depth0pt \kern2pt}%
	\def\@fs@post{\kern2pt\hrule\relax}%
	\def\@fs@mid{\kern2pt\hrule\kern2pt}%
	\let\@fs@iftopcapt\iftrue}
\makeatother

\usepackage{xargs} 
\usepackage{graphicx}
\usepackage{soul}
\usepackage{pgfgantt}
\usepackage{pdflscape}
\usepackage{lipsum}
\usepackage{pstricks}
\usepackage{import}
\usepackage{xcolor}
\definecolor{midgrey}{RGB}{100, 100, 100}
\definecolor{darkgreen}{RGB}{0, 100, 50}
\definecolor{plum}{RGB}{78, 25, 96}
\usepackage{layouts}
\usepackage{array,booktabs,tabularx}
\usepackage{ragged2e}
\usepackage{nccmath}

\usepackage[colorinlistoftodos,prependcaption,textsize=small]{todonotes}
\usepackage{varwidth}
\usepackage{multirow}
\usepackage{multicol}
\usepackage{siunitx}
\usepackage{amsmath,amssymb}
\usepackage{mathtools,stackengine}
\usepackage{dsfont}
\DeclareFontFamily{U}{mathc}{}
\DeclareFontShape{U}{mathc}{m}{it}%
{<->s*[1.03] mathc10}{}
\DeclareMathAlphabet{\mathscr}{U}{mathc}{m}{it}

\usepackage{pgf}

\usepackage{algorithm}
\usepackage[noend]{algpseudocode}
\usepackage{scalerel}
\usepackage{stackengine}

\usepackage{url}

\usepackage{caption}

\usepackage[toc, sort=def, acronym, style = long]{glossaries}

\makeatletter
\edef\@figdir{_pgfcache}

\catcode`\#=11
\catcode`\|=6
\newcommand{\@basicpgfpreamble}[1]{%
    \unexpanded{%
        \documentclass{standalone}^^J
        \usepackage{pgf}^^J
        \RequirePackage{luatex85}^^J
        \let\oldpgfimage\pgfimage^^J
        \renewcommand{\pgfimage}[2][]{\oldpgfimage[#1]{|1/#2}}^^J
    }%
}
\catcode`\#=6
\catcode`\|=12

\let\@pgfpreamble\@basicpgfpreamble

\newcommand{\setpgfpreamble}[1]{%
    \renewcommand{\@pgfpreamble}[1]{\@basicpgfpreamble{##1}\unexpanded{#1}}
}

\newcounter{@pgfcounter}
\newwrite\@pgfout
\newread\@pgfin

\newcommand{\importpgf}[3][]{%
    \IfFileExists{#2/#3}{}{\errmessage{importpgf: File #2/#3 not found}}%
    \edef\@figfile{\jobname-\the@pgfcounter}%
    \providecommand{\@writetempfile}{}%
    \renewcommand{\@writetempfile}[1]%
    {%
        \immediate\openout\@pgfout=##1%
        \immediate\write\@pgfout{\@pgfpreamble{#2}}%
        \immediate\write\@pgfout{\string\begin{document}}%
        \immediate\openin\@pgfin=#2/#3%
        \begingroup\endlinechar=-1%
            \loop\unless\ifeof\@pgfin%
                \readline\@pgfin to \@fileline%
                \ifx\@fileline\@empty\else%
                    \immediate\write\@pgfout{\@fileline}%
                \fi%
            \repeat%
        \endgroup%
        \immediate\closein\@pgfin%
        \immediate\write\@pgfout{\string\end{document}}%
        \immediate\closeout\@pgfout%
    }%
    \def\@compile%
    {%
        \immediate\write18{lualatex -interaction=batchmode -output-directory="\@figdir" \@figdir/\@figfile.tex}%
    }%
    \IfFileExists{\@figdir/\@figfile.pdf}%
    {%
        \@writetempfile{\@figdir/tmp.tex}%
        \edef\@hashold{\pdfmdfivesum file {\@figdir/\@figfile.tex}}%
        \edef\@hashnew{\pdfmdfivesum file {\@figdir/tmp.tex}}%
        \ifnum\pdfstrcmp{\@hashold}{\@hashnew}=0%
            \relax%
        \else%
            \@writetempfile{\@figdir/\@figfile.tex}%
            \@compile%
        \fi%
    }%
    {%
        \@writetempfile{\@figdir/\@figfile.tex}%
        \@compile%
    }%
    \IfFileExists{\@figdir/\@figfile.pdf}%
    {\includegraphics[#1]{\@figdir/\@figfile.pdf}}%
    {\errmessage{Error during compilation of figure #2/#3}}%
    \stepcounter{@pgfcounter}%
}
\makeatother

\renewcommand{\thefootnote}{\fnsymbol{footnote}}

\newcommandx{\unsure}[2][1=]{\todo[inline,linecolor=red,backgroundcolor=red!25,bordercolor=red,#1]{#2}}
\newcommandx{\info}[2][1=]{\todo[inline,linecolor=OliveGreen,backgroundcolor=OliveGreen!25,bordercolor=OliveGreen,#1]{#2}}
\newcommandx{\improvement}[2][1=]{\todo[inline,linecolor=Plum,backgroundcolor=Plum!25,bordercolor=Plum,#1]{#2}}
\newcommandx{\thiswillnotshow}[2][1=]{\todo[disable,#1]{#2}}
\newcommand{\rework}[1]{%
	\todo[color=yellow!25,inline, caption={}]{\normalsize {#1}}%
}
\newcommand{\idea}[1]{%
	\todo[color=blue!25,inline, caption={}]{\normalsize {#1}}%
}

\newcommand{\important}[1]{%
	\todo[color=red!25,inline, caption={}]{\normalsize {#1}}%
}
\newcommand{\evaluation}[1]{%
	\todo[color=green!25,inline, caption={}]{\normalsize {#1}}%
}
\newcommand{\futurework}[1]{%
	\todo[color=plum!25,inline, caption={}]{\normalsize {#1}}%
}

\graphicspath{{./pictures/}{./figures/}{./../pics/}}

\newcommand{\figurepath}{plots}
\newcommand{\figurepathnew}{../../../Programming/ma_sp/paper_eval/figures/}

\newcommand{\tablepath}{../../../Programming/ma_sp/paper_eval/tables}

\newcolumntype{C}{>{\Centering}X}
\newcolumntype{L}{>{\RaggedRight}X}
\newcolumntype{R}{>{\RaggedLeft}X}

\newcommand\equalhat{\mathrel{\stackon[1.5pt]{=}{\stretchto{%
				\scalerel*[\widthof{=}]{\wedge}{\rule{1ex}{3ex}}}{0.5ex}}}}
\DeclareMathOperator{\E}{\mathbb{E}}
\DeclareMathOperator*{\argmax}{argmax}
\DeclareMathOperator*{\argmin}{argmin} 
\DeclareMathOperator*{\infinum}{inf} 
\newcommand{\Rho}{\mathrm{P}}
\newcommand{\Astar}{$\text{A}^* $ }

\renewcommand{\Return}{\gls{Return}}
\newcommand{\Returnmean}{\gls{Returnmean}}
\newcommand{\reward}{\gls{reward}}
\newcommand{\Reward}{\gls{Reward}}
\newcommand{\SThresh}{\gls{SThresh}}

\newcommand{\Real}[1]{\ensuremath{\mathbb{R}^{#1}}} 
\newcommand{\Naturals}[1]{\ensuremath{\mathbb{N}^{#1}}} 
\newcommand{\DistrUniform}{\ensuremath{\mathcal{U}}} 
\newcommand{\BigO}{\ensuremath{\mathcal{O}}} 
\newacronym{RL}{RL}{reinforcement rearning}
\glsadd{RL}

\newacronym{cpu}{CPU}{Central Processing Unit}
\glsadd{cpu}

\newacronym{UCB}{UCB}{Upper Confidence Bound}

\newcommand{\est}[1]{\ensuremath{s^{#1}}} 
\newcommand{\EST}[1]{\ensuremath{\mathcal{S}^{#1}_\text{env}}}
\newcommand{\ost}[2]{\ensuremath{o_{#2}^{#1}}} 
\newcommand{\OST}[2]{\ensuremath{\mathcal{O}_{#2}^{#1}}}
\renewcommand{\ast}[2]{\ensuremath{s_{#2}^{#1}}} 
\newcommand{\AST}[2]{\ensuremath{\mathcal{S}_{#2}^{#1}}}
\newcommand{\PSP}{\ensuremath{\mathcal{\Pi}}} 
\newcommand{\bst}[3]{\ensuremath{b_{#2#3}^{#1}}} 
\newcommand{\BST}[3]{\ensuremath{\prescript{#3}{}{\mathcal{B}_{#2}^{#1}}}} 
\newcommand{\BSTInt}[3]{\ensuremath{\prescript{#3}{}{B_{#2}^{#1}}}} 
\newcommand{\BSTSize}{\ensuremath{N_B}} 
\newcommand{\hbst}[2]{\ensuremath{\theta_{#2}^{#1}}}
\newcommand{\HBST}[2]{\ensuremath{\Theta_{#2}^{#1}}}
\newcommand{\ist}[2]{\ensuremath{i_{#2}^{#1}}} 
\newcommand{\IST}[2]{\ensuremath{\mathcal{I}_{#2}^{#1}}}
\newcommand{\HIST}[2]{\ensuremath{\prescript{#2}{}{\mathbb{H}}_{I}^{#1}}}

\newcommand{\pst}[1]{\ensuremath{s^{#1}_\text{prior}}} 
\newcommand{\PST}[1]{\ensuremath{\mathcal{S}^{#1}_\text{prior}}} %

\newcommand{\numhyp}{\ensuremath{K}} 
\newcommand{\hypidx}{\ensuremath{k}} 
\newcommand{\hyp}[1]{\ensuremath{\theta_{#1}}} 
\newcommand{\HYP}[1]{\ensuremath{\Theta_{#1}}} 

\newcommand{\sh}[1]{\ensuremath{H^{#1}_\text{s}}}
\newcommand{\SH}{\ensuremath{\mathcal{H}_\text{s}}}

\newcommand{\oh}[1]{\ensuremath{H^{#1}_\text{o}}}
\newcommand{\OH}{\ensuremath{\mathcal{H}_\text{o}}}

\newcommand{\of}[1]{\ensuremath{F^{#1}_\text{o}}}
\newcommand{\OF}{\ensuremath{\mathcal{F}_\text{o}}}

\renewcommand{\a}[3]{\ensuremath{a_{#2#3}^{#1}}}
\newcommand{\A}[2]{\ensuremath{A_{#2}^{#1}}}

\newcommand{\p}[1]{\ensuremath{\pi_{#1}}}
\newcommand{\ptrue}[1]{\ensuremath{\pi_{#1}}}
\newcommand{\phyp}{\ensuremath{\pi^{*}}}
\newcommand{\pest}[1]{\ensuremath{\widehat{\pi}_{#1}}}

\newcommand{\TEnv}{\ensuremath{T_\text{env}}}
\newcommand{\TPrior}{\ensuremath{T_\text{prior}}}
\newcommand{\TPriorSpace}{\ensuremath{\mathcal{T}_\text{prior}}}
\newcommand{\THypo}{\ensuremath{T_\text{hypo}}}

\newcommand{\PEnv}{\ensuremath{\mathbb{P}_\text{env}}}
\newcommand{\PScen}{\ensuremath{\mathbb{P}_\text{s}}}
\newcommand{\PPrior}{\ensuremath{\mathbb{P}_\text{prior}}}
\newcommand{\scen}{\ensuremath{\mathcal{s}}}
\newcommand{\Scen}{\ensuremath{\mathcal{S}}}
\newcommand{\goaldef}{\ensuremath{\mathcal{g}}}
\newcommand{\Goaldef}{\ensuremath{\mathcal{G}}}
\newcommand{\miles}{\ensuremath{m}}
\newcommand{\PFuture}{\ensuremath{\mathbb{P}}}

\newcommand{\Smeas}{\ensuremath{f}}
\newcommand{\SmeasEnv}{\ensuremath{\Smeas_\text{envelope}}}
\newcommand{\SmeasCol}{\ensuremath{\Smeas_\text{collision}}}
\newcommand{\GRisklevel}{\ensuremath{\beta}}
\newcommand{\SRisklevel}{\ensuremath{\beta}}
\newcommand{\SRisk}{\ensuremath{\rho_\text{scenario}}}
\newcommand{\DRisk}{\ensuremath{\rho_\text{decision}}}
\newcommand{\HRisk}{\ensuremath{\rho_\text{history}}}

\newcommand{\SRisklevelEst}{\ensuremath{\beta_\text{est}}}
\newcommand{\Efficiency}{\ensuremath{\epsilon}}

\newcommand{\Runtime}{\ensuremath{\mathsf{T}}}

\newcommand{\TScen}{\ensuremath{\mathsf{T}_\text{sc}}}
\newcommand{\TDriven}{\ensuremath{\mathsf{T}_\text{d}}}
\newcommand{\TViolate}{\ensuremath{\mathsf{T}_\text{v}}}
\newcommand{\TFuture}{\ensuremath{\mathsf{T}}}
\newcommand{\TStep}{\ensuremath{\tau}}
\newcommand{\TPlan}{\ensuremath{\mathsf{T}_\text{Plan}}}

\newcommand{\egoidx}{\ensuremath{i}}
\newcommand{\otheridx}{\ensuremath{j}}
\newcommand{\egonumact}[2]{\ensuremath{k_\egoidx}}
\newcommand{\numotheragents}{\ensuremath{N_\otheridx}}

\newcommand{\discount}{\ensuremath{\gamma}}

\newcommand{\desiredgap}[2]{\ensuremath{d^{#1}_{#2}}}
\newcommand{\desiredgapleft}[1]{\ensuremath{d_{l,#1}}}
\newcommand{\desiredgapright}[1]{\ensuremath{d_{r,#1}}}

\newcommand{\Unknown}{\ensuremath{\raisebox{.5pt}{\textcircled{\raisebox{-.9pt} {?}}}}}

\newcommand{\idmdesiredvelocity}{\ensuremath{v_\text{desired}}}
\newcommand{\idmminimumspacing}{\ensuremath{s_\text{min}}}
\newcommand{\idmdesiredheadway}{\ensuremath{T_\text{desired}}}
\newcommand{\idmaccfactor}{\ensuremath{\dot{v}_\text{factor}}}
\newcommand{\idmaccmin}{\ensuremath{\dot{v}_\text{min}}}
\newcommand{\idmaccmax}{\ensuremath{\dot{v}_\text{max}}}
\newcommand{\idmcomftbrake}{\ensuremath{\dot{v}_\text{comft}}}

\newcommand{\acccoolness}{\ensuremath{C_\text{coolness}}}

\newglossaryentry{Return}{%
	name=\ensuremath{R},
	description={Random Return Variable distributed according to return distribution $R\sim Z$}
}

\newglossaryentry{Returnmean}{%
	name=\ensuremath{\overline{R}},
	description={Mean of return $\overline{R} = \E_{R\sim Z}[R]$}
}

\newglossaryentry{reward}{%
	name=\ensuremath{\mathscr{r}},
	description={single reward value given for transition from $s$ to $s^\prime$ executing action $a$}
}

\newglossaryentry{Reward}{%
	name=\ensuremath{\mathcal{R}},
	description={Reward distribution for transition from $s$ to $s^\prime$ executing action $a$}
}

\newglossaryentry{SThresh}{%
	name=\ensuremath{\lambda_\text{safety}~},
	description={Defines the amount of safety the decision making algorithm shall achieve}
}

\newglossaryentry{EnvStateT}{%
	name=\est{t},
	description={The environment state at time $t$, $\est{t}=(\ast{t}{0}, \ast{t}{1},\ldots,\ast{t}{n})$}
}

\newglossaryentry{EnvStateSpace}{%
	name=\EST{},
	description={The space of environment states}
}

\newglossaryentry{PriorEnvStateT}{%
	name=\pst{t},
	description={The prior environment state at time $t$, $\pst{t}=(\ast{t}{0}, \ast{t}{1},\ldots,\ast{t}{n})$}
}

\newglossaryentry{PriorEnvStateSpace}{%
	name=\PST{},
	description={The space of prior environment states shall be defined as a superset of the environment state space $\PST{} \supseteq \EST{}$. }
}

\newglossaryentry{AgentStateT}{%
	name=\ast{t}{i},
	description={The full state of agent $i$ at time $t$, $\ast{t}{i} =(\ost{t}{i},\ist{t}{i}, \bst{t}{i} )$}
}

\newglossaryentry{AgentStateSpace}{%
	name=\AST{}{i},
	description={The state space of agent $i$, $\AST{}{i} = \OST{}{i} \times \IST{}{} \times \BST{}{i}{}$ }
}

\newglossaryentry{ObservableAgentStateT}{%
	name=\ost{t}{i},
	description={The observable state of agent $i$ at time $t$, in our case the physical state}
}

\newglossaryentry{ObservableAgentStateSpace}{%
	name=\OST{}{i},
	description={The space of observable states of agent $i$}
}

\newglossaryentry{ObservableStateT}{%
	name=\ost{t}{},
	description={The observable state of the environment at time $t$, in our case the physical states of all observable agents}
}

\newglossaryentry{BehaviorAgentStateT}{%
	name=\bst{t}{i}{},
	description={The behavior state of agent $i$ at time $t$  }
}

\newglossaryentry{BehaviorStateT}{%
	name=\bst{t}{}{},
	description={The environment behavior state of all observable agents at time $t$,$\bst{t}{}{}=(\bst{t}{0}{},\bst{t}{1}{},\ldots,\bst{t}{n}{})$   }
}

\newglossaryentry{BehaviorStateSpace}{%
	name=\BST{}{i}{},
	description={The space of behavior states}
}

\newglossaryentry{IntentionAgentStateT}{%
	name=\ist{t}{i},
	description={The intention state of agent $i$ at time $t$ }
}

\newglossaryentry{IntentionStateT}{%
	name=\ist{t}{},
	description={The environment intention state of all observable agents at time $t$, $\ist{t}{}=(\ist{t}{0},\ist{t}{1},\ldots,\ist{t}{n})$ }
}

\newglossaryentry{IntentionStateSpace}{%
	name=\IST{}{},
	description={The space of intentions}
}

\newglossaryentry{StateHistoryT}{%
	name=\sh{t},
	description={The state action history up to time $t$ starting from the initial observation state \ost{0}, $H^t=(\est{0},\a{0}{}{},\est{1},\a{1}{}{},\ldots,\est{t})$}
}

\newglossaryentry{StateHistorySpace}{%
	name=\SH,
	description={The space of state action histories.}
}

\newglossaryentry{ObservationHistoryT}{%
	name=\oh{t},
	description={The observation action history up to time $t$ starting from the initial observation state \ost{0}, $H^t=(\ost{0},\a{0}{}{},\ost{1},\a{1}{}{},\ldots,\ost{t})$}
}

\newglossaryentry{ObservationHistorySpace}{%
	name=\OH,
	description={The space of observation action histories }
}

\newglossaryentry{AgentAction}{%
	name=\a{t}{i}{k},
	description={The $k$th action of agent $i$ at time $t$}
}

\newglossaryentry{JointAction}{%
	name=\a{t}{}{},
	description={The joint action $\a{t}{}{}=(\a{t}{1}{},\a{t}{2}{},\ldots,\a{t}{n}{})$ selected by the agents at time $t$}
}

\newglossaryentry{ActionSet}{%
	name=\A{t}{i},
	description={The set of possible actions of agent $i$ at time $t$, $\A{t}{i}=(\a{t}{i}{1},\a{t}{i}{2},\ldots,\a{t}{i}{k_i})$ where $k_i^t$ is the number of actions of agent $i$ at time $t$ }
}

\newglossaryentry{AgentPolicy}{%
	name=\p{i},
	description={The stochastic policy of agent $i$ mapping action observations histories \oh{t} to action \a{t}{}{} with $\p{i} : \OH \times \A{t}{i} \rightarrow [0,1] $ }
}

\newglossaryentry{WorldTransition}{%
	name=\TEnv,
	description={The stochastic environment transition function maps the ego-agents action \a{t}{0}{} to the next state: $\TEnv: \A{}{0} \times \OH{} \times \EST{} \rightarrow [0,1]$     }
}

\newglossaryentry{EnvironmentDistribution}{%
	name=\PEnv,
	description={The environment distribution models what environment states \est{0} and transition functions \TEnv are encountered initially at the start of a scenario: $\est{0}, \TEnv \sim \PEnv$    }
}

\newglossaryentry{PriorEnvironmentDistribution}{%
	name=\PPrior,
	description={The prior environment distribution models what prior environment states \pst{0} and transition functions \TPrior are encountered initially at the start of a scenario: $\pst{0},\TPrior \sim \PPrior$    }
}

\newacronym[plural=sbg,firstplural=Stochastic Bayesian Games (SBGs)]{sbg}{SBG}{Stochastic Bayesian Game}
\newacronym[plural=MDPs,firstplural=Markov Decision Processes (MDPs)]{mdp}{MDP}{Markov Decision Process}
\newacronym{rmdp}{RMDP}{Robust Markov Decision Process}
\newacronym[plural=RSBGs,firstplural=Robust Stochastic Bayesian Games (RSBGs)]{rsbg}{RSBG}{Robust Stochastic Bayesian Game}
\newacronym[plural=POMSPs,firstplural=Partially Observable Markov Decision Processes (POMDPs)]{pomdp}{POMDP}{Partially Observable Markov Decision Process}
\newacronym[plural=CC-POMDPs,firstplural=Cost-Constrained Partially Observable Markov Decision Processes (CC-POMDPs)]{ccpomdp}{CC-POMDP}{Cost-Constrained Partially Observable Markov Decision Process}
\newacronym[plural=RC-RSBGs,firstplural=Risk-Constrained Robust Stochastic Bayesian Games (RC-RSBGs)]{rcrsbg}{RC-RSBG}{Risk-Constrained Robust Stochastic Bayesian Game}
\newacronym[plural=MA-MDPs,firstplural=Multi-Agent Markov Decision Processes (MA-MDPs)]{mamdp}{MA-MDP}{Multi-Agent Markov Decision Process}

\newacronym{hba}{HBA}{Harsanyi Bellman Ad Hoc}
\newacronym{bamdp}{BAMDP}{Bayesian-adaptive \gls{mdp}}

\newacronym{mcts}{MCTS}{Monte Carlo Tree Search}
\newacronym{mpc}{MPC}{Model Predictive Control}
\newacronym{mamcts}{MA-MCTS}{Multi-Agent Monte Carlo Tree Search}

\newtheorem{theorem}{Theorem}[section]
\newtheorem{corollary}{Corollary}[theorem]
\newtheorem{lemma}[theorem]{Lemma}
\newtheorem{definition}{Definition}[section]
\newtheorem{proposition}{Theorem}[section]

\setpgfpreamble{
\usepackage{xargs} 
\usepackage{graphicx}
\usepackage{soul}
\usepackage{pgfgantt}
\usepackage{pdflscape}
\usepackage{lipsum}
\usepackage{pstricks}
\usepackage{import}
\usepackage{xcolor}
\definecolor{midgrey}{RGB}{100, 100, 100}
\definecolor{darkgreen}{RGB}{0, 100, 50}
\definecolor{plum}{RGB}{78, 25, 96}
\usepackage{layouts}
\usepackage{array,booktabs,tabularx}
\usepackage{ragged2e}
\usepackage{nccmath}

\usepackage[colorinlistoftodos,prependcaption,textsize=small]{todonotes}
\usepackage{varwidth}
\usepackage{multirow}
\usepackage{multicol}
\usepackage{siunitx}
\sisetup{load-configurations = abbreviations}

\usepackage{amsmath,amssymb}
\usepackage{mathtools,stackengine}
\usepackage{dsfont}
\DeclareFontFamily{U}{mathc}{}
\DeclareFontShape{U}{mathc}{m}{it}%
{<->s*[1.03] mathc10}{}
\DeclareMathAlphabet{\mathscr}{U}{mathc}{m}{it}

\usepackage{pgf}

\usepackage{algorithm}
\usepackage[noend]{algpseudocode}
\usepackage{scalerel}
\usepackage{stackengine}

\usepackage{url}

\usepackage{caption}

\usepackage[toc, sort=def, acronym, style = long]{glossaries}

\renewcommand{\thefootnote}{\fnsymbol{footnote}}

\newcommandx{\unsure}[2][1=]{\todo[inline,linecolor=red,backgroundcolor=red!25,bordercolor=red,#1]{#2}}
\newcommandx{\info}[2][1=]{\todo[inline,linecolor=OliveGreen,backgroundcolor=OliveGreen!25,bordercolor=OliveGreen,#1]{#2}}
\newcommandx{\improvement}[2][1=]{\todo[inline,linecolor=Plum,backgroundcolor=Plum!25,bordercolor=Plum,#1]{#2}}
\newcommandx{\thiswillnotshow}[2][1=]{\todo[disable,#1]{#2}}
\newcommand{\rework}[1]{%
	\todo[color=yellow!25,inline, caption={}]{\normalsize {#1}}%
}
\newcommand{\idea}[1]{%
	\todo[color=blue!25,inline, caption={}]{\normalsize {#1}}%
}

\newcommand{\important}[1]{%
	\todo[color=red!25,inline, caption={}]{\normalsize {#1}}%
}
\newcommand{\evaluation}[1]{%
	\todo[color=green!25,inline, caption={}]{\normalsize {#1}}%
}
\newcommand{\futurework}[1]{%
	\todo[color=plum!25,inline, caption={}]{\normalsize {#1}}%
}

\graphicspath{{./pictures/}{./figures/}{./../pics/}}

\newcommand{\figurepath}{plots}
\newcommand{\figurepathnew}{../../../Programming/ma_sp/paper_eval/figures/}

\newcommand{\tablepath}{../../../Programming/ma_sp/paper_eval/tables}

\newcolumntype{C}{>{\Centering}X}
\newcolumntype{L}{>{\RaggedRight}X}
\newcolumntype{R}{>{\RaggedLeft}X}

\newcommand\equalhat{\mathrel{\stackon[1.5pt]{=}{\stretchto{%
				\scalerel*[\widthof{=}]{\wedge}{\rule{1ex}{3ex}}}{0.5ex}}}}
\DeclareMathOperator{\E}{\mathbb{E}}
\DeclareMathOperator*{\argmax}{argmax}
\DeclareMathOperator*{\argmin}{argmin} 
\DeclareMathOperator*{\infinum}{inf} 
\newcommand{\Rho}{\mathrm{P}}
\newcommand{\Astar}{$\text{A}^* $ }

\renewcommand{\Return}{\gls{Return}}
\newcommand{\Returnmean}{\gls{Returnmean}}
\newcommand{\reward}{\gls{reward}}
\newcommand{\Reward}{\gls{Reward}}
\newcommand{\SThresh}{\gls{SThresh}}

\newcommand{\Real}[1]{\ensuremath{\mathbb{R}^{#1}}} 
\newcommand{\Naturals}[1]{\ensuremath{\mathbb{N}^{#1}}} 
\newcommand{\DistrUniform}{\ensuremath{\mathcal{U}}} 
\newcommand{\BigO}{\ensuremath{\mathcal{O}}} 
		
\renewcommand{\egoidx}{\ensuremath{i}}
\renewcommand{\otheridx}{\ensuremath{j}}
}

\renewcommand{\egoidx}{\ensuremath{i}}
\renewcommand{\otheridx}{\ensuremath{j}}

\newcommand{\plotpath}{./plots}
\graphicspath{{./pictures/}{./figures/}{./pics/}{\plotpath}}

\begin{document}
	
\title{Risk-Constrained Interactive Safety under Behavior Uncertainty for Autonomous Driving}

\author{Julian Bernhard$^{1}$ and Alois Knoll$^{2}$%
	\thanks{$^{1}$Julian Bernhard is with fortiss GmbH, An-Institut Technische Universit\"{a}t M\"{u}nchen, Munich, Germany}%
	\thanks{$^{2}$Alois Knoll is with Chair of Robotics, Artificial Intelligence and Real-time Systems, Technische Universit\"{a}t M\"{u}nchen, Munich, Germany}
}

\maketitle

\begin{abstract}
Balancing safety and efficiency when planning in dense traffic is challenging. Interactive behavior planners incorporate prediction uncertainty and interactivity inherent to these traffic situations. Yet, their use of single-objective optimality impedes interpretability of the resulting safety goal. Safety envelopes which restrict the allowed planning region yield interpretable safety under the presence of behavior uncertainty, yet, they sacrifice efficiency in dense traffic due to conservative driving. Studies show that humans balance safety and efficiency in dense traffic by \emph{accepting a probabilistic risk} of violating the safety envelope.
 In this work, we adopt this safety objective for interactive planning. Specifically, we formalize this safety objective, present the Risk-Constrained Robust Stochastic Bayesian Game modeling interactive decisions satisfying a maximum risk of violating a safety envelope under uncertainty of other traffic participants' behavior and solve it using our variant of Multi-Agent Monte Carlo Tree Search. We demonstrate in simulation that our approach outperforms baselines approaches, and by reaching the specified violation risk level over driven simulation time, provides an interpretable and tunable safety objective for interactive planning. 
\end{abstract}

\section{Introduction}
	
    Behavior planners for autonomous vehicles must be able to solve dense driving situations in close interaction with humans thereby being confronted with a variety of behavioral variations and limited knowledge about the true human driving behavior. 
    
    Interactive behavior planners became more and more advanced to plan in such situations. Popular approaches to predict other participants use cooperative \cite{wang_enabling_2020, lenz_tactical_2016, kurzer_decentralized_2018-1} or probabilistic models \cite{hubmann_belief_2018, lu_safe_2020}. The authors presented a behavior space approach in \cite{bernhard_robust_2020} to deal with the variety of human behavior variations in interactive planning. Yet, existing approaches leave open how to specify a meaningful safety objective under the presence of behavior uncertainty. They use single-objective optimality criteria which are not interpretable with respect to a meaningful safety goal, e.g a maximum collision probability~$P_\text{col}$.
    
     Restricting the ego motion to stay within a safety envelope, e.g. defined using reachability analysis \cite{pek_using_2020, leung_infusing_2020, yu_risk_2020} or other forms of safe distance measures \cite{shalev-shwartz_formal_2017, pierson_learning_2019} circumvents the problem of prediction model inaccuracy. However, since interactions are neglected, this approach becomes problematic in crowded traffic where it leads to conservative driving or the freezing vehicle symptom.      

	\begin{figure}[t]																				
		\def\svgwidth{\columnwidth}
		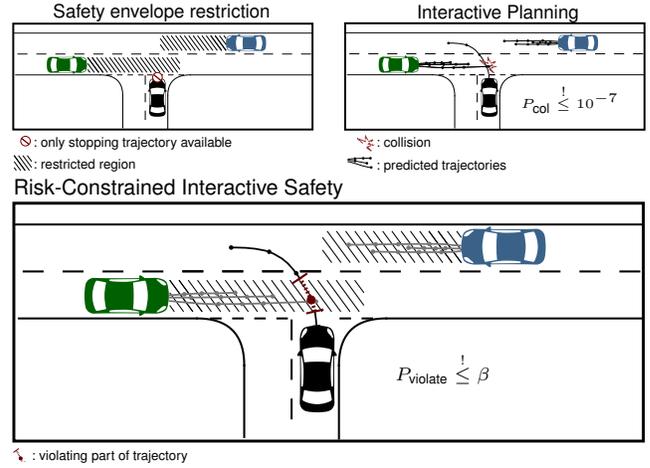 \vspace{-5mm}
		\caption{Restricting the ego motion to stay within a safety envelope fosters the freezing vehicle symptom in dense traffic scenarios. Objective functions of interactive planners employable in dense traffic are not interpretable with respect to the collision probability $P_\text{col}$. Therefore, we transfer a human-related safety objective for dense traffic to interactive planning. Our planner generates a policy which satisfies a specified maximum risk level of violating a safety envelope $\SRisklevel$. } \label{fig:top_level}
		\vspace{-4mm}
	\end{figure}
    It seems that humans follow a different safety goal than realized by existing behavior planners. Evaluation over real-world driving data showed that humans violate safe distance measures with a certain percentage \emph{averaged over driven time} \cite{esterle_formalizing_2020} with increasing violations occurring in dense traffic during rush hours \cite{ pek_verifying_2017}. It seems that the safety objective of human drivers balances safety and efficiency in a comprehensible way. 
    
    In this work, we therefore adopt this safety objective for interactive planning. Our approach generates a policy which satisfies a maximum risk of violating a safety envelope under uncertainty of other traffic participants' behavior. Specifically, we contribute
    \begin{itemize}
    	\item a formalization of this safety objective,
    	\item the \gls{rcrsbg} modeling risk-constrained interactive decisions under behavior uncertainty, 
    	\item a \gls{mamcts} planner to solve the \gls{rcrsbg},
    	\item a simulative analysis showing that our approach outperforms baselines approaches while reaching the specified violation risk level averaged over driven time. 
    	\end{itemize}

Fig. \ref{fig:top_level} visualizes our contribution. We start with related work. Next, we formalize the problem and present our method followed by the experimental evaluation.

\section{Related Work}
We present related work on risk metrics and levels, interactive and risk-constrained planning.

\subsection{Risk-metrics and levels}
There exist probabilistic and non-probabilistic risk definitions.
The latter defines a metric and an accompanying risk threshold. Data-related metrics use human driving data to parameterize distance functions, e.g to fit Gaussian metrics \cite{pierson_learning_2019} or human safety corridors from feedback in simulation \cite{wei_risk-based_2019}. Kinematics-based metrics model physics-based risk, e.g time-to-collision \cite{iberraken_safe_2018} or potential fields to model braking forces \cite{akagi_stochastic_2015} and are closely related to safety envelope definitions \cite{shalev-shwartz_formal_2017, pierson_learning_2019} whereby a kinematic risk level of zero corresponds to completely staying within the safety envelope. Non-probabilistic risk is straightforward to define and interpret. Yet, it does not reveal information about the probability of occurrence of a harmful event and neglects uncertainty information, e.g. beliefs about the behavior of other participants.

In the functional safety sense, risk is probabilistic and defined as the combination of probability of occurrence of harm and the severity of that harm \cite{hruschka_uncertainty-adaptive_2019}. Existing probabilistic risk definitions often consider collision as harmful events in risk-based planning approaches \cite{huang_hybrid_2018, huang_hybrid_2018, yu_risk_2020, yu_occlusion-aware_2019, hruschka_uncertainty-adaptive_2019} using the state probability, the probability of spatial overlap at discrete times \cite{philipp_analytic_2019}. Value-based probabilistic risk definitions arose in finance and are applied to robotics and autonomous driving \cite{majumdar_how_2017, bernhard_addressing_2019, ge_risk-aware_2019}. Defining risk based on collisions is infeasible considering the vast amount of samples required to approximate human collision probabilities $P_{col}{\approx}10^{-7}$. In contrast to previous work, we propose a notion of risk coupled to how human drivers might balance safety and efficiency in interactive situations based on the risk of violating a safety envelope. By formalizing risk as event probability, with the harmful event occurring over a period of time \cite{philipp_analytic_2019}, correspondence between \emph{specified risk level and observed risk} is achieved.

\subsection{Interactive Planning}
In \cite{bernhard_bark_2020}, \citeauthor{bernhard_bark_2020} define the behavior of an autonomous vehicle as \emph{its desired future} sequence of physical states encoding the agent's strategy to reach a short-term goal, e.g changing lane. A behavior planner creates a behavior trajectory and passes it to a controller. In this work, we focus on modeling of risk induced by unknown behaviors of other traffic participants and drop localization and execution uncertainties. Interactive planning algorithms incorporate potential reactions of other participants into their plan to successfully navigate in congested traffic \cite{schwarting_planning_2018}. We briefly summarize existing approaches to predict other participants behavior within an interactive planning process.

\emph{Cooperative approaches} assume that participants act according to a global cost function in a \gls{mamdp} \cite{kurzer_decentralized_2018-1, wang_enabling_2020, lenz_tactical_2016} and allow for explainable parameterization of the model. However, the assumption that traffic follows a globally optimal solution neglects the uncertainty inherent to human interactions. \emph{Markov approaches} represent participants' behavior via environment transitions based on the current observable state in a \gls{mdp} \cite{bernhard_addressing_2019, bouton_reinforcement_2018}. Though, \glsplural{mdp} model uncertainty in state transitions, they neglect information from past states, facilitating inefficient and unsafe decisions. In constrast, \emph{Belief-based approaches} use observations from past states to gather information about the true behavior of other participants, modeled as \gls{pomdp} \cite{lu_safe_2020, hubmann_belief_2018, bai_intention-aware_2015, bouton_belief_2017} or Bayesian game \cite{bernhard_robust_2020} to improve efficiency in dense traffic scenarios.

Existing interactive planners employ a \emph{single-objective} optimality criterion with manual or data-based cost tuning to avoid collisions \cite{wang_enabling_2020, lenz_tactical_2016, kurzer_decentralized_2018-1, bernhard_addressing_2019, lu_safe_2020, hubmann_belief_2018, bai_intention-aware_2015}. 
Compared to previous work, we employ a \emph{multi-objective} optimality criterion to model risk over behavior uncertainty. For this, we combine our decision model, the \gls{rsbg} presented in \cite{bernhard_robust_2020} with the  \glsfirst{ccpomdp}~\mbox{\cite{isom_piecewise_2008, lee_monte-carlo_2018}}. 

\subsection{Risk-Constrained Planning}
Planning under non-probabilistic risk definitions equals finding an ego trajectory within the space spanned by risk metric and level, e.g by using graph search on a discretized state space \cite{pierson_learning_2019} or \gls{mpc} \cite{zhou_gap_2019}. 
Probabilistic collision risk is used in \cite{ge_risk-aware_2019} to model lane changing as conditional value at risk \gls{mdp}, in \cite{huang_hybrid_2018} to model on ramp merging as \glsentryshort{ccpomdp} and in \cite{hruschka_uncertainty-adaptive_2019} to incorporate various uncertainties into an MPC algorithm. Related to our risk definition, \citet{muller_risk_2019} constrain the risk of violating the safe distance. 
Prescribed related work in risk-constrained planning employs long-term, i.e maneuver-based prediction of other participants neglecting interactions.
Interactive risk-constrained planning using reinforcement learning has been investigated in \cite{bernhard_addressing_2019} using conditional value at risk, and in \cite{bouton_reinforcement_2018} using a discretized state space. 

The presented approaches allow for specification of a risk constraint, yet, reveal missing correspondence between \emph{specified and observed risk} in the experimental evaluation. 
In this work, we demonstrate that with our variant of \gls{mamcts} the observed risk in simulation corresponds to the specified constraint yielding an interpretable and tunable safety objective for interactive planning.

\section{Problem Formulation} \label{sec:problem_formulation}
Evaluation over real-world driving data showed that humans violate safe distance measures with a certain percentage~\cite{esterle_formalizing_2020, pek_verifying_2017}. It seems that humans
\begin{enumerate}
	\item \textbf{stay within} their safety envelope, e.g. spanned by the current safe distance, \emph{in most cases} to avoid unsafe behavior due to modeling uncertainty, yet
	\item \textbf{accept the risk\footnote{Modeling severity of envelope violations is considered future work.}} of violating the safety envelope. They behave such that a safety envelope is violated with not more than probability \SRisklevel~ under consideration of prediction uncertainty.   
\end{enumerate}
By adjusting \SRisklevel~ humans tune safety versus efficiency to avoid conservative driving in congested, rush hour traffic~\cite{pek_verifying_2017}. Based on these considerations, we informally define a human-related safety objective for interactive behavior planning as: "The ego vehicle behaves such that the percentage of time the safety envelope is violated is smaller than a given threshold". Next, we formalize this safety goal. 

We model the traffic environment in a game-theoretic manner~\cite{albrecht_belief_2016}. It consists of $\numotheragents$ other agents, i.e traffic participants, each observing a joint environment state $\ost{t}{}=(\ost{t}{1},\ost{t}{2},\ldots,\ost{t}{N}) \in \OST{}{}$ with \emph{physical} dynamic and static properties, e.g. participant positions and velocities, map information etc. Agents choose next actions based on their policy $\a{t}{\otheridx}{}\sim \ptrue{\otheridx}(\a{t}{j}{}|\ost{t}{}, \Unknown)$ defined over a continuous action space \A{}{j}{}, e.g. a 2-dimensional space bounding maximum longitudinal acceleration and steering angle. We control a single agent $\egoidx$, the autonomous vehicle, which selects actions from a discrete set of actions \A{}{\egoidx} according to $\a{}{\egoidx}{} \sim\p{\egoidx}$. It reasons about the behavior of the other agents $\otheridx$. Agent $i$ knows the action space and can observe past actions of the other agents. The true policy \ptrue{\otheridx} and hidden inputs \Unknown~ are \emph{unknown} to \egoidx.  Deterministic transitions to the next environment state, $\ost{t+1}{}=T(\ost{t}{},\a{t}{}{})$, result from the agents' joint action $\a{t}{}{} = (\a{t}{\egoidx}{}, \a{t}{-\egoidx}{})$ and their kinematic models, e.g. single track, applied for action duration $\tau_a$ to \ost{t}{}. An index $-\egoidx$ indicates joint action of all agents except agent \egoidx. The process continues until some terminal criterion applies, e.g. collision or goal reached. 
Using this environment model we define a violation risk.

\begin{definition} \textbf{Violation risk:}
	Given the current environment state $\ost{t}{} \in \OST{}{}$, behavior policies $\p{\egoidx}$ and $\p{\otheridx}$, a safety violation indicator $\Smeas: \OST{}{} \rightarrow \{0, 1\}$ indicating a safety violation in observation $\ost{\prime}{} \in\OST{}{}$, then the violation risk is defined as:
	\begin{flalign}
\label{eq:scenario_risk_def}	\rho(\ost{t}{}, \p{\egoidx}, \p{\otheridx}, \Smeas) {=} \E_{ \of{} {\sim} \PFuture^{\ost{t}{}, \p{\egoidx}, \p{\otheridx}}}\bigg[ \frac{\sum_{z=1}^{z=|\of{}| - 1} \Smeas(\of{}(z)) \cdot \tau_a}{|\of{}|\cdot \tau_a} \bigg]               
\end{flalign}
\end{definition} 
\vspace{2mm}
The expectation is defined over the distribution $\PFuture^{\ost{t}{}, \p{\egoidx}, \p{\otheridx}}$ over future observation sequences $\of{}=(\ost{t}{},\ost{t+\tau_a}{}, \ost{t+2\tau_a}{},\ldots)$ starting from current environment state $\ost{t}{}$. It is influenced by ego and other agents policies. The upper sum represents the violation duration within the observation sequence \of{} with $|\of{}|$ being the length of the sequence and $\of{}(z)$ giving the $z$-th observation within the sequence. The lower term is the total duration of the sequence. The fraction of these terms yields the percentage of time the safety envelope is violated for one sequence. A sequence ends in a terminal state. The temporal resolution of this fraction is determined by the action duration $\tau_a$. For fixed ego policy $\p{\egoidx}$, the expectation provides the \emph{time-based} violation risk under unknown behavior of other participants $\p{\otheridx}$.

Using equation \ref{eq:scenario_risk_def}, we formalize risk-constrained interactive safety against behavior uncertainty.  
\begin{definition} \textbf{Risk-constrained safety:}
	Given an indicator function for safety envelope violations \SmeasEnv, the behavior planner generates a goal-directed policy \p{\egoidx} in the current environment state $\ost{t}{}$ under unknown behavior $\p{\otheridx}$ of other participants which achieves a safety envelope violation risk lower than a specified allowed risk level $\SRisklevel$:
	\begin{equation}
 \label{eq:risk_threshold}
	\rho(\ost{t}{}, \p{\egoidx}, \p{\otheridx}, \SmeasEnv) \widehat{=} \rho_\text{env}(\ost{t}{}, \p{\egoidx}, \p{\otheridx}) \overset{!}{\leq} \SRisklevel \\
	\end{equation}
\end{definition}
	\begin{figure}[t]																			\vspace{2mm}	
	\def\svgwidth{\columnwidth}
	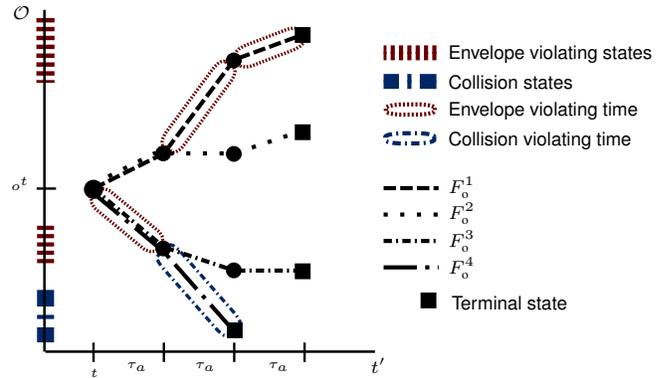 \vspace{-3mm}
	\caption{Example for envelope violation and collision risk calculation: Four future observation sequences \of{1-4} starting from state \ost{t}{} are sampled with probabilities  $\PFuture(\of{1-3}) = 0.3$ and $\PFuture(\of{4}) = 0.1$, with \of{4} ending due to a terminal collision state. Since sequence \of{1} shows 2 and sequences \of{3-4} show 1 envelope violation, we obtain $\rho_\text{env}(\cdot)=0.3\cdot\frac{2\tau_a}{3\tau_a}+0.3\cdot\frac{0\tau_a}{3\tau_a}+0.3\cdot\frac{1\tau_a}{3\tau_a}+0.1\cdot\frac{1\tau_a}{2\tau_a}=0.35$. The only sequence showing a collision violation is \of{4} giving $\rho_\text{col}(\cdot)=0.3\cdot\frac{0\tau_a}{3\tau_a}+0.3\cdot\frac{0\tau_a}{3\tau_a}+0.3\cdot\frac{0\tau_a}{3\tau_a}+0.1\cdot\frac{1\tau_a}{2\tau_a}=0.05$.} \label{fig:state_tree}
	\vspace{-7mm}
\end{figure}
\vspace{2mm}
Our risk formulation is independent of the goal formalism, e.g. based on rewards, and the actual safety envelope definition e.g. being based on safe distance measures or reachability analysis. Since, the collision risk is difficult to interpret and infeasible to calculate exactly, we require it to be close to zero,~$\rho(\ost{t}{}, \p{\egoidx}, \p{\otheridx}, \SmeasCol) \widehat{=} \rho_\text{col}(\ost{t}{}, \p{\egoidx}, \p{\otheridx}) \overset{!}{\approx} 0$ with \SmeasCol~denoting a collision indicator. Fig.~\ref{fig:state_tree} provides an example of envelope and collision risk calculations. 

To satisfy our problem definition, the planner must correctly approximate the observation sequence distribution under unknown behavior of other agents and optimize a multi-objective criterion integrating the safety envelope and collision risk constraints and the goal-directed optimality criterion. In the next section, we propose a game-theoretic model, the \gls{rcrsbg} and a variant of \gls{mamcts} to solve the defined problem.

\section{Risk-Constrained Interactive Planning} \label{sec:rcrsbg}
We present a game-theoretic model, the \glsfirst{rcrsbg}, which combines
\begin{itemize}
	\item the \gls{rsbg} presented by \citet{bernhard_robust_2020} to approximate the observation sequence distribution in eq.~\ref{eq:scenario_risk_def} given unknown behavior of other participants $\p{\otheridx}$
	\item and the \gls{ccpomdp} to incorporate risk constraints,
	\item and adapt a \gls{mamcts} planner to solve the \gls{rcrsbg}.
\end{itemize}
We detail each of these aspects in the following. 
\subsection{Robust Stochastic Bayesian Game (RSBG)} \label{sec:rsbg}
In \citet{bernhard_robust_2020}, the authors presented the \glsfirst{rsbg}, a game-theoretic model for interactive planning to cover  the \emph{physically} feasible \emph{continuous} variations inherent to human behavior considering both inter-driver and intra-driver variabilities \cite{uhrmacher_multi-agent_2009}. The model uses a predefined set of hypothetical behavior types  $\hbst{\hypidx}{} \in \HBST{}{}$ and behavior hypotheses $\a{t}{j}{} \sim \p{\hbst{\hypidx}{}}(\a{t}{j}{}|\oh{t}{})$. To predict the others' behavior one tracks the posterior beliefs $\text{Pr}(\hbst{\hypidx}{}|\oh{t}{}, \otheridx)$ over hypothesized types for each agent based on the action-observation history $\oh{t}{} \in \OH$. 

To obtain behavior types and hypothesis, it defines a hypothetical policy 
\begin{equation}
\phyp: \OH \times  \BST{t}{j}{}  \rightarrow \A{}{\otheridx}
\end{equation}
for a specific task with $\bst{t}{\otheridx}{}\in\BST{t}{j}{}$ being \otheridx th agent's  behavior state at time $t$, $\BST{t}{j}{} \subset \Real{\BSTSize}$ its behavior space of dimension \BSTSize. A behavior state \bst{t}{\otheridx}{} is a \emph{physically interpretable} quantity describing \otheridx th continuous behavior variations. The other agents' behavior spaces $\BST{t}{j}{}$ and their current behavior states \bst{t}{\otheridx}{} are unknown. By using the property of physical interpretability of \bst{t}{\otheridx}{}, an expert can define a hypothesized behavior space \BST{}{}{}, comprising the individual behavior spaces $\BST{}{j}{}$ ($\BST{}{j}{} \subset \BST{}{}{}$), by looking at the physically realistic situations. For instance, it is straightforward to define the physical boundaries of a behavior state modeling the desired gap between vehicles \otheridx~ and \egoidx~ at the time point of merging onto another lane with the one-dimensional behavior space $\BST{}{}{}=\{\bst{}{}{}| \bst{}{}{}\in [0, d_\text{max}]\}$ where $d_\text{max}$ is the maximum sensor range. 
The approach then uses a partitioning of the full behavior space $\BST{}{}{}=\BST{1}{}{} \cup \BST{2}{}{} \cup \ldots \cup \BST{\numhyp}{}{}, \, \forall l\neq k \BST{l}{}{} \cap \BST{k}{}{} = \emptyset$. This yields \numhyp~hypothesis $\p{\hbst{\hypidx}{}}$ by adopting a uniform distribution over behavior states within behavior space partitions $\BST{k}{}{}$, respectively.    

\floatstyle{spaceruled}
\restylefloat{algorithm}
\begin{algorithm*}[t]
	\begin{minipage}[t]{0.4\textwidth}
		\setlength{\textfloatsep}{0pt}
		\begin{algorithmic}
	\scriptsize
	\Function{Search}{$\ost{t}{}{}, \text{Pr}(\hbst{\hypidx}{}|\oh{t}{}, \cdot)$}
	\State $\lambda \gets$ \textsc{RandomInit()}
	\Repeat
	\For{$\otheridx=1\ldots\numotheragents$}
	\State $\hbst{\prime}{\otheridx} \sim \text{Pr}(\hbst{\hypidx}{}|\oh{t}{}, \otheridx)$
	\EndFor
	\State \textsc{Simulate($\langle\ost{t}{}{}\rangle,\hbst{\prime}{\otheridx} \forall \otheridx$, 1)}
	\State $\a{}{\egoidx}{} \sim\,$\textsc{EgoActionSelection($\langle\ost{t}{}{}\rangle, 0$)}
	\State $\lambda_\text{env} \gets \lambda_\text{env} + \alpha_n[\rho_\text{env}(\langle\ost{t}{}{}\rangle, \a{}{\egoidx}{}) - \SRisklevel]$
	\State $\lambda_\text{col} \gets \lambda_\text{col} + \alpha_n[\rho_\text{col}(\langle\ost{t}{}{}\rangle, \a{}{\egoidx}{}) - 0]$
	\State Clip $\lambda_\text{env},\, \lambda_\text{col}$ to range $[0, 1]$
	\Until{\textsc{MaxIterations()}}
	\State \textbf{return} \textsc{EgoActionSelection($\langle\ost{t}{}{}\rangle, 0$)} 
	\EndFunction
\end{algorithmic}

		\setlength{\textfloatsep}{0pt}
		\begin{algorithmic}
	\scriptsize 
	\Function{OtherActionSelection}{$\langle \oh{}\rangle, j, \hbst{}{j}$}
	\If{\textsc{ProgressiveWidening}()} 
		\State \textbf{return} $\a{}{}{j} \gets \p{\hbst{}{j}}(\a{}{\otheridx}{}| \oh{}{})$
	\Else 
	\State \textbf{return} $\text{argmax}_{\a{}{}{}} Q_{\overline{C}}(\langle\oh{}{}\rangle, \a{}{\otheridx}{}, j)$
	\EndIf
	\EndFunction
\end{algorithmic}

		\setlength{\textfloatsep}{0pt}
		\begin{algorithmic}
	\scriptsize 
	\Function{EgoActionSelection}{$\langle\oh{}\rangle, \kappa$}
	\State \parbox[b][1cm][b]{0.3\textwidth}{\begin{equation*}\begin{split}
		\scriptstyle
		Q^{\oplus}_\lambda(\langle\oh{}\rangle, & \a{}{}{}) \gets \scriptstyle Q_{\Return}(\langle\oh{}\rangle, \a{}{}{}) - \lambda_\text{env} \cdot \rho_\text{env}(\langle\oh{}\rangle, \a{}{}{}) \\ & \scriptstyle - \lambda_\text{col} \cdot \rho_\text{col}(\langle\oh{}\rangle, \a{}{}{}) 
		 + \kappa\sqrt{\log N(\langle\oh{}\rangle)/N(\langle\oh{}\rangle, \a{}{}{}, \egoidx)}
	\end{split}\end{equation*}}
	\State $\a{*}{}{} \gets \text{arg max}_{\a{}{}{}} Q^{\oplus}_\lambda(\langle\oh{}\rangle, \a{}{}{})$
	\State $\A{*}{} \gets$ add other actions to \a{*}{}{} to consider exploration differences
	\State $\p{\egoidx} \gets$ Solve linear program with $ \A{*}{}$ to obtain stochastic policy
	\State \textbf{return} $\a{}{\egoidx}{} \sim \p{\egoidx}$ 
	\EndFunction
\end{algorithmic}

	\end{minipage}
	\begin{minipage}[t]{0.6\textwidth}
		\begin{algorithmic}
	\scriptsize
	\Function{Simulate}{$\langle\oh{}{}\rangle, \hbst{\prime}{\otheridx} \forall \otheridx, d$}
	\If{$d  > d_\text{max}$ \textbf{or} \textsc{IsTerminal}($\langle\oh{}{}\rangle$)}
	\State \textbf{return} [0, 0, 0 ,0, 0]
	\EndIf
	\If{\textsc{FirstNodeVisit(}$\langle\oh{}{}\rangle$)}
		\State \textbf{return} \textsc{RandomRollout}($\langle\oh{}{}\rangle, \hbst{\prime}{\otheridx} \forall \otheridx, d$)
	\EndIf
	\State $\a{}{\egoidx}{} \gets $\textsc{EgoActionSelection}($\langle\oh{}{}\rangle, \kappa$)
	\For{$l=1\ldots\numotheragents$}
		\State $\a{}{\otheridx}{l} \gets $\textsc{OtherActionSelection}($\langle\oh{}{}\rangle, l, \hbst{\prime}{l}$)
	\EndFor
	\State $\tau_\text{predict} \gets d\cdot\tau_a$
	\State $(\ost{\prime}{},r) \gets\,$\textsc{EnvironmentMove}($\oh{}{},(\a{}{\egoidx}{},\a{}{\otheridx}{}), \tau_\text{predict}$)
	\State  $[R^\prime, T_\text{env}^\prime, T_\text{col}^\prime, T_\text{tot}^\prime, \overline{C}^\prime] \gets $\textsc{Simulate}($\langle\oh{}{}, (\a{}{\egoidx}{},\a{}{\otheridx}{}), \ost{\prime}{}\rangle, \hbst{\prime}{\otheridx} \forall \otheridx, d+1$)
	\State \parbox[b][1cm][b]{0.6\textwidth}{\begin{equation*}\begin{split} [R, T_\text{env}, T_\text{col}, T_\text{tot}] \gets &
				[r + \discount \cdot R^\prime, T_\text{env}^\prime + \SmeasEnv(\ost{\prime}{})\cdot\tau_\text{predict}, \\ &T_\text{col}^\prime + \SmeasCol(\ost{\prime}{})\cdot\tau_\text{predict}, T_\text{tot}^\prime +  \tau_\text{predict}]	\end{split}\end{equation*}} \vspace{-2mm}
	\State $N(\langle\oh{}{}\rangle) \gets N(\langle\oh{}{}\rangle) + 1$
	\State $N(\langle\oh{}{}\rangle, \a{}{\egoidx}{}, \egoidx) \gets N(\langle\oh{}{}\rangle, \a{}{\egoidx}{}, \egoidx) + 1$
	\State $Q_R(\langle\oh{}{}\rangle, \a{}{\egoidx}{}) \gets Q_R(\langle\oh{}{}\rangle, \a{}{\egoidx}{}) + (R-Q_R(\langle\oh{}{}\rangle, \a{}{\egoidx}{}))/N(\langle\oh{}{}\rangle, \a{}{\egoidx}{}, \egoidx)$ 
	\State $\rho_\text{env}(\langle\oh{}{}\rangle, \a{}{\egoidx}{}) \gets \rho_\text{env}(\langle\oh{}{}\rangle, \a{}{\egoidx}{}) + (T_\text{env}/T_\text{tot}-\rho_\text{env}(\langle\oh{}{}\rangle, \a{}{\egoidx}{}))/N(\langle\oh{}{}\rangle, \a{}{\egoidx}{}, \egoidx)$
	\State $\rho_\text{col}(\langle\oh{}{}\rangle, \a{}{\egoidx}{}) \gets \rho_\text{col}(\langle\oh{}{}\rangle, \a{}{\egoidx}{}) + (T_\text{col}/T_\text{tot}-\rho_\text{col}(\langle\oh{}{}\rangle, \a{}{\egoidx}{}))/N(\langle\oh{}{}\rangle, \a{}{\egoidx}{}, \egoidx)$
	\For{$l=1\ldots\numotheragents$}
	\State $N(\langle\oh{}{}\rangle, \a{}{\otheridx}{l}, l) \gets N(\langle\oh{}{}\rangle, \a{}{\otheridx}{l}, l) + 1$
	\State $\overline{C} \gets \SmeasEnv(\ost{\prime}{}) + \SmeasCol(\ost{\prime}{}) + \gamma\cdot \overline{C}^\prime$
	\State $Q_{\overline{C}}(\langle\oh{}{}\rangle, \a{}{\otheridx}{l}, l) \gets Q_{\overline{C}}(\langle\oh{}{}\rangle, \a{}{\otheridx}{l}) + (\overline{C}-Q_{\overline{C}}(\langle\oh{}{}\rangle, \a{}{\otheridx}{l}))/N(\langle\oh{}{}\rangle, \a{}{\otheridx}{l}, \otheridx)$ 
	\EndFor
	\State \textbf{return} $[R, T_\text{env}, T_\text{col}, T_\text{tot}, \overline{C}]$
	\EndFunction
\end{algorithmic}

	\end{minipage}
	\caption{Multi-Agent Monte Carlo Tree Search for the Risk-Constrained Robust Stochastic Bayesian Game} \label{alg:ccrsbg}
\end{algorithm*}

The \gls{rsbg} computes an optimal ego policy integrating the current posterior type belief $\text{Pr}(\hbst{\hypidx}{}|\oh{t}{}, \otheridx)$ according to $\p{\egoidx} = \max_{\p{\egoidx}} Q^{\p{\egoidx}}(\oh{t}{})$, where $Q^{\p{\egoidx}}(\oh{t}{})=$
\begin{equation} \label{eq:rsbg_optimality}
\begin{split}
\E_{\p{\egoidx}, (\hbst{\hypidx}{\otheridx}, \ldots)_{\forall j} \sim \text{Pr}(\cdot|\oh{t}{}, \otheridx)}&\bigg[
\sum_{t^\prime=t}^{\infty} \discount^t r(\ost{t^\prime}{},\a{t^\prime}{}{})\bigg]\\ \medmath{\text{with}\,
\a{t^\prime}{}{} }&\medmath{= (\a{t^\prime}{\egoidx}{}, \a{t^\prime}{-\egoidx}{}),} \\
\medmath{\a{t^\prime}{\egoidx}{}} & \medmath{\sim \p{\egoidx}(\cdot|\ost{t^\prime}{}), }\\
\medmath{\a{t^\prime}{-\egoidx}{}} &\medmath{= (\a{t^\prime}{\otheridx}{}, \ldots)_{\forall j} \sim \p{\hbst{\hypidx}{\otheridx}}}
\end{split}
\end{equation} 
is the expected cumulative reward of agent $\egoidx$ in state \ost{t}{} and history \oh{t}{}. Future rewards $r(\ost{t}{},\a{t}{}{})$ are discounted by $\discount$. We denote independent sampling from $P$ and sample concatenation for all other agents \otheridx~with $(\cdot,\ldots)_{\forall \otheridx} \sim P$.

\subsection{Risk-Constrained Robust Stochastic Bayesian Game} \label{sec:rcrsbg}
Next, we integrate the risk constraints for safety envelope violation and collision into the \gls{rsbg}. For this, we approximate the unknown behavior of other agents \p{\otheridx} in the constraint eq.~\ref{eq:scenario_risk_def} using a mixture distribution $\pest{\otheridx}$ combining hypotheses and posterior beliefs for each other agent \otheridx: 
\begin{equation}
\pest{\otheridx}(\a{}{\otheridx}{}|\oh{t}{})= \sum_{\forall k} \text{Pr}(\hbst{\hypidx}{}|\oh{t}{}, \otheridx)\cdot \p{\hbst{\hypidx}{}}(\a{}{\otheridx}{}|\oh{t}{})
\end{equation}   
Optimality  of \gls{rcrsbg}s is then defined by extending eq.~\ref{eq:rsbg_optimality} with the risk constraints.
\begin{definition} \textbf{Optimality of \gls{rcrsbg}s}
The optimal ego policy $\p{\egoidx}$ of the \gls{rcrsbg} maximizes the expected cumulative reward $Q^{\p{\egoidx}}(\oh{t}{})$ defined in eq.~\ref{eq:rsbg_optimality} subject to 
\begin{equation} \label{eq:rcrsbg_optimality}
\begin{split} 
\rho_\text{env}(\ost{t}{}, \p{\egoidx}, \pest{\otheridx}) &\overset{!}{\leq} \SRisklevel \\
\rho_\text{col}(\ost{t}{}, \p{\egoidx}, \pest{\otheridx}) &\overset{!}{\approx} 0
\end{split} 
\end{equation} 
\end{definition}
\vspace{2mm}
Next, we present a \glsfirst{mamcts} approach which solves the \gls{rcrsbg}. 

\subsection{Monte Carlo Tree Search for the \gls{rcrsbg}}
Our \gls{mamcts} approach is shown in Alg.~\ref{alg:ccrsbg}. To solve the \gls{rsbg}, as presented in \cite{bernhard_robust_2020}, we use: 
\begin{itemize}
	\item Stage-wise action selection: Similar to \cite{kurzer_decentralized_2018-1, lenz_tactical_2016} agents select actions independently in stages. Their joint action yields the next environment state. In constrast to previous work \cite{kurzer_decentralized_2018-1, lenz_tactical_2016}, we define separate selection mechanisms for ego and other agents' in \textsc{EgoActionSelection} and \textsc{OtherActionSelection} each returning actions \a{}{\egoidx}{} or \a{}{\otheridx}{} at stage nodes $\langle \oh{}\rangle$. Ego actions are discrete. Other agents' actions are continuous and their selection strategy represents the agents' behavior types. 
	\item Hypothesis-belief-sampling: At the beginning of each search iteration, we sample a hypothesis for each other agent $j$ from the posterior belief $\hbst{\prime}{j} \sim \text{Pr}(\hbst{\hypidx}{}|\oh{t}{}, \otheridx)$ and use it within the function \textsc{OtherActionSelection} for selection, expansion and roll-out steps.
\end{itemize}

To extend the \gls{mamcts} approach to solve the \gls{rcrsbg}, we develop an ego action selection mechanism \textsc{EgoActionSelection} based on solutions to \glsplural{ccpomdp}.  The \gls{pomdp} is a well known single-agent framework to model sequential decisions under partially observable environment states.  An optimal policy of a \glsfirst{ccpomdp} \cite{isom_piecewise_2008, lee_monte-carlo_2018} not only maximizes the expected cumulative reward but also constraints $M$ expected cumulative costs $Q_{C_m} \leq \widehat{c}_m,\, m\in M$. 

\citet{lee_monte-carlo_2018} use \gls{mcts} to solve \gls{ccpomdp}s by reformulating the problem as an \emph{unconstrained} POMDP. They introduce Lagrange multipliers and express the optimal value function as $ Q^*_{\lambda} = Q_{R} - \sum_{\forall m}\lambda^*_m \cdot Q_{C_m}$ for optimal $\lambda^*_m$.
A solution is found by updating the Lagrange multipliers iteratively at the beginning of each \gls{mcts} iteration using a gradient estimate $\Delta \lambda_m \sim Q_{C_m} - \widehat{c}_m, \, \forall m$. 

We adapt their approach to solve the \gls{rcrsbg}. Yet, instead of cumulative costs, we maintain ego envelope violation and collision action-risks $\rho_\text{env}(\langle \oh{}\rangle, \a{}{\egoidx}{})$ and $\rho_\text{col}(\langle \oh{}\rangle, \a{}{\egoidx}{})$ in each stage node.  
For this, we separately backpropagate the violation durations for safety envelope  $T_\text{env}$ and collision $T_\text{col}$, and total time $T_\text{tot}$ occurred with this iteration's selection, expansion and rollout step in the \textsc{Simulate}() function. These terms correspond to the upper and lower term of the ratio defined in eq.~\ref{eq:scenario_risk_def} and can be used to update $\rho_\text{env}(\langle \oh{}\rangle)$ and $\rho_\text{col}(\langle \oh{}\rangle)$ in each iteration.  Return estimates are updated as usual. Prediction time $\tau_\text{predict}$ increases with search depth $d$.

In the search method, we perform a gradient update of the Lagrange multipliers $\lambda_\text{env}$ and $\lambda_\text{col}$ using the root node's risks estimates $\rho_\text{env}(\langle \ost{t}{}{}\rangle, \a{}{\egoidx}{})$ and $\rho_\text{col}(\langle \ost{t}{}{}\rangle, \a{}{\egoidx}{})$ and desired risk constraints~\SRisklevel~ and 0 with decreasing step size \mbox{$\alpha_n \sim 1/$\textsc{NumIterations}()}. In \textsc{EgoActionSelection}, we calculate the combined action-value based on the estimated Lagrange multipliers. As proposed in \cite{lee_monte-carlo_2018}, we account for inaccuracies in return and risk estimates and form a set of equal-valued actions \A{*}{} maximizing the action-value by accepting a tolerance based on action selection counts. An optimal policy of a \gls{ccpomdp} is generally stochastic. In our case, this requires
\begin{equation} \label{eq:stochastic_policy}
\begin{split}
  \sum_{ \a{}{\egoidx}{} \in \A{*}{}} \p{\egoidx}(\a{}{\egoidx}{}|\ost{t}{}{}) \cdot\rho_\text{env}(\langle\ost{t}{}{}\rangle, \a{}{\egoidx}{}) &\overset{!}{\leq} \SRisklevel \\
    \sum_{ \a{}{\egoidx}{} \in \A{*}{}} \p{\egoidx}(\a{}{\egoidx}{}|\ost{t}{}{})\cdot \rho_\text{col}(\langle\ost{t}{}{}\rangle, \a{}{\egoidx}{}) &\overset{!}{=} 0
  \end{split}
 \end{equation}
We solve the linear program defined in \cite{lee_monte-carlo_2018} to obtain a stochastic policy \p{\egoidx} satisfying eq.~\ref{eq:stochastic_policy}, \ref{eq:rsbg_optimality} and \ref{eq:rcrsbg_optimality}.

We apply worst-case action selection in combination with progressive widening for other agents in \textsc{OtherActionSelection}. Other agents maintain separate \emph{combined} ego cost estimates $Q_{\overline{C}}(\langle\oh{}{}\rangle, \a{}{\otheridx}{}, j)$ during back-propagation for their own selected actions. Progressive widening ensures that new actions are explored. In the other case, other agents select actions \emph{maximizing} the combined envelope violation and collision cost of the ego agent. As shown in \cite{bernhard_robust_2020}, this concept reduces sample complexity when other agents select actions from a continuous action space. In the case of a constrained setting, it additionally improves convergence by better exploring joint actions which violate the given risk constraints.

\section{Experiment}
In our experiment, we evaluate quantitatively over many driven scenarios with simulated behavior uncertainty
\begin{itemize}
	\item if the presented planning approach respects the specified risk level \SRisklevel,
	\item how the risk parameter \SRisklevel~balances safety and efficiency,
	\item and how our approach compares against baseline interactive planners.
\end{itemize}
We also qualitatively analyze the computed policy and posterior beliefs in specific scenarios. 
\setlength{\dbltextfloatsep}{2pt}
\begin{figure*}[t!]
	\vspace{2mm}
	\centering
	\includegraphics[width=1.0\textwidth]{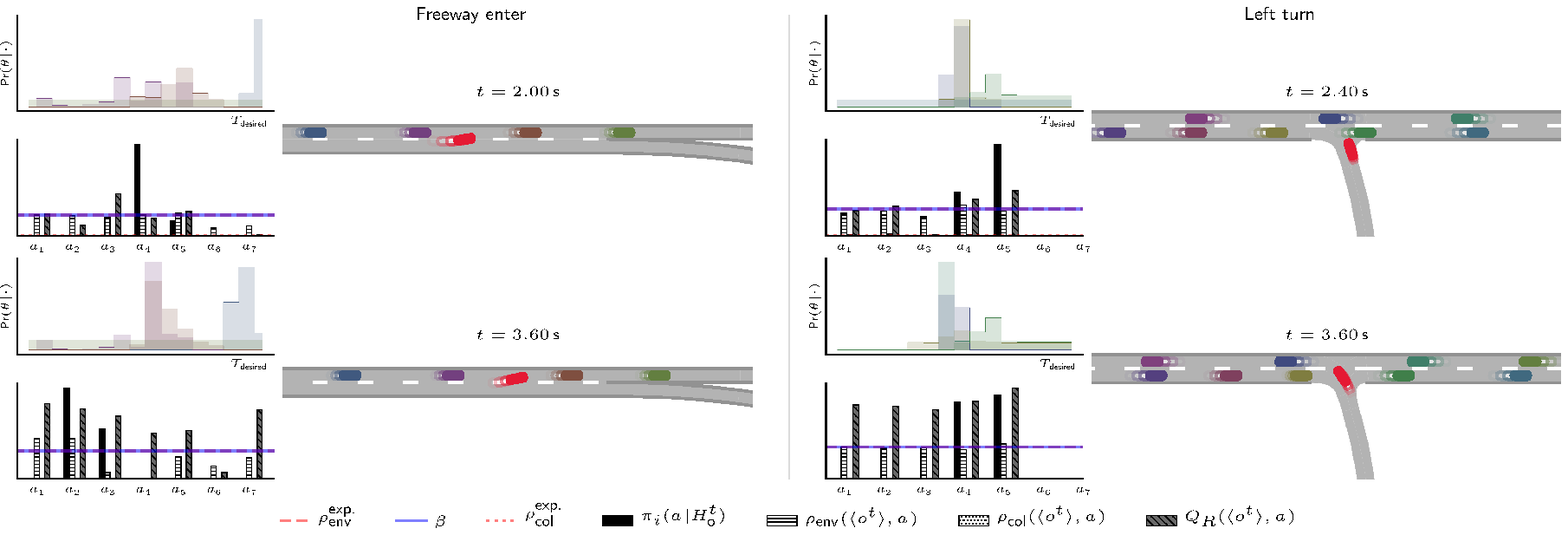}
	\vspace{-6mm}
	\caption{Scenario analysis at two time steps with posterior beliefs $\text{Pr}(\hbst{\hypidx}{}|\oh{t}{}, \otheridx)$ for four nearest agents in respective color. The planned stochastic policy $\p{\egoidx}$ balances action-risk estimates, $\rho_\text{env}(\langle\ost{t}{}{}\rangle, \a{}{\egoidx}{})$ and $\rho_\text{col}(\langle\ost{t}{}{}\rangle, \a{}{\egoidx}{})$ yielding an expected envelope risk $\rho_\text{env}^\text{exp.}$ fulfilling the risk constraint $\SRisklevel=0.2$ while the expected planned collision risk $\rho_\text{col}^\text{exp.}$ is close to zero and higher returns $Q_R(\langle\oh{}{}\rangle, \a{}{\egoidx}{})$ are preferred.}
	\label{fig:qualitative_results} 
\end{figure*}

\subsection{Scenarios}
We use the OpenSource behavior benchmarking environment BARK \cite{bernhard_bark_2020} for simulating two dense traffic scenarios: 
\begin{itemize}
	\item Freeway enter: In a double merge, the ego vehicle wants to enter the freeway on the left occupied lane. 
	\item Left turn: The ego vehicle wants to turn left from a side into a main road having to cross two occupied lanes.
	\end{itemize}
We uniformly sample initial starting conditions, e.g. the distances between vehicles from $[\SI{15}{\meter}, \SI{30}{\meter}]$ and velocities from $[\SI{30/3.6}{\meter\per\second}, \SI{50/3.6}{\meter\per\second}]\,\text{}$. Fig.~\ref{fig:qualitative_results} provides examples of the scenarios. A scenario successfully terminates when the ego vehicle is close to and oriented towards the goal while the velocity is within the sampling bounds.

\subsection{Behavior Simulation \& Space}
The Intelligent Driver Model (IDM) \cite{uhrmacher_multi-agent_2009} defines both the simulated unknown behavior \p{\otheridx}~of other participants and hypothetical policy \phyp~used for hypotheses design. For simulation, we define a 5-dimensional true behavior space \BST{*}{5D}{} over the IDM parameters and uniformly draw unknown boundaries of behavior variations $[\bst{l}{}{j, \text{min}}, \bst{l}{}{j,\text{max}}],\, l\in \{1,\ldots,5\}$ for each agent and trial ($\BST{}{\otheridx}{}\subseteq \BST{*}{5D}{}$). We introduce the parameters minimum and maximum boundary widths $\Delta_\text{min}$/$\Delta_\text{max}$ to specify minimum and maximum time-dependent variations of behavior states in simulation. This avoids unrealistic large variations of behavior parameters. 

For hypotheses design, we employ a 1D-behavior space, since, as shown in \cite{bernhard_robust_2020}, lower-dimensional behavior spaces can capture the uncertainty occurring with the 5-dimensional true behavior space. Tab.~\ref{tab:behavior_space} depicts both the simulated 5D and hypothesized 1-D behavior space used in our experiment.

We simulate the other agents by sampling a new behavior state \bst{t}{\otheridx}{} at every time step from the unknown boundaries of behavior variations \BST{}{\otheridx}{} and then use the IDM model with this parameters to choose their actions. Fixing the random seeds for all sampling operations ensures equal conditions for all evaluations.

\begin{table}[b]
	\tiny
	\vspace{-4mm}
	\centering
	\begin{tabular}{p{0.15\columnwidth}*{2}{>{\centering\arraybackslash}p{0.15\columnwidth}}*{1}{>{\centering\arraybackslash}p{0.09\columnwidth}}} 
    \toprule
    &  \multicolumn{2}{c@{}}{\BST{*}{5D}{} } &  \BST{}{1D, \text{Head.}}{} \\
    \midrule 
    Param \bst{l}{}{} & [\bst{l}{}{\text{min}},\bst{l}{}{\text{max}}] & $\Delta_\text{min}$/$\Delta_\text{max}$ & [\bst{l}{}{\text{min}},\bst{l}{}{\text{max}}]   \\ 
    \midrule 
   \idmdesiredvelocity\, [m/s] 		& [8.0, 14.0] & 0.1 / 0.1   &  11.0   \\
   \idmdesiredheadway\, [s] 		& [0.5, 2.0] & 0.1 / 0.3   &   [0.0, 4.0] \\
   \idmminimumspacing\, [1]   				& [2.0, 2.5] & 0.1 / 0.3    & 2.25   \\ 
   \idmaccfactor\, [m/$\text{s}^2$] & [1.5, 2.0] & 0.1 / 0.3  &  1.75   \\
  \idmcomftbrake\, [m/$\text{s}^2$] & [1.7, 2.0] & 0.1 / 0.3  &   1.85   \\ 
    \bottomrule
    \end{tabular}
	\caption{Boundaries of the simulated true behavior spaces \BST{*}{5D}{} and the hypothesized behavior space \BST{}{1D, \text{Head.}}{} for IDM parameters desired velocity \idmdesiredvelocity, desired time headway \idmdesiredheadway, minimum spacing \idmminimumspacing, acceleration factor \idmaccfactor, comfortable braking \idmcomftbrake.} \label{tab:behavior_space}
\end{table}
\setlength{\dbltextfloatsep}{1pt}
\begin{figure*}[t!]
	\centering
	\vspace{2mm}
	\includegraphics[width=1.0\textwidth]{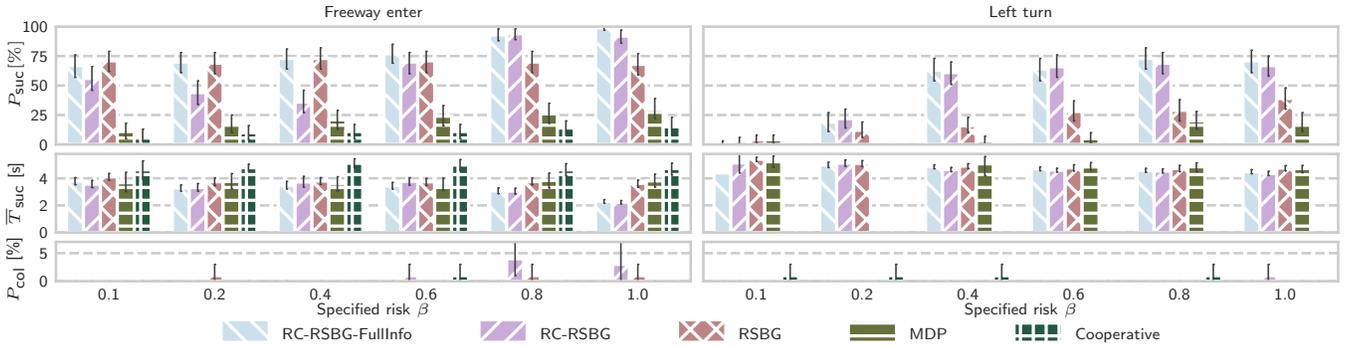}
	\vspace{-5.5mm}
	\caption{Performance metrics for \textbf{RC-RSBG} planner and baselines for increasing envelope violation risk levels $\SRisklevel$.  }
	\label{fig:barplot_results} 
\end{figure*}

\begin{figure}[b]
	\vspace{-5mm}
	\centering
	\includegraphics[width=1.0\columnwidth]{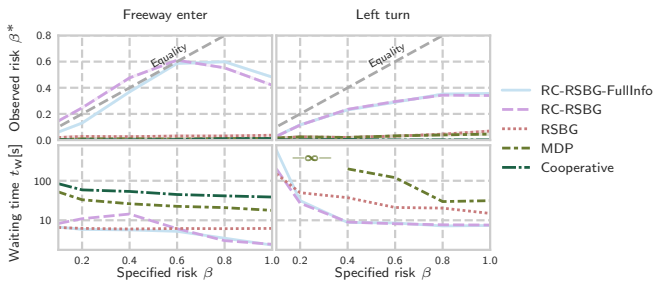}
	\vspace{-4mm}
	\caption{Analysis of observed envelope violation risk $\SRisklevel^{*}$ and expected scenario waiting time $t_\text{w}$ for different risk levels \SRisklevel.}
	\vspace{-3mm}
	\label{fig:risk_mean} 
\end{figure}

\subsection{Safety Violation Indicators}
\citet{rizaldi_formalising_2017} propose a model for longitudinal safe distances between vehicles which we use to define the indicator \SmeasEnv~to return a violation if the ego vehicle violates the safe distance to other vehicles.  The response time of other vehicles is \SI{1}{\second}. For the ego vehicle, we use \SI{0.5}{\second} in freeway and, due to higher traffic density, \SI{1}{\second} in left turn. The deceleration limit is \SI{-5}{\meter\per\second\squared} as in the IDM model.

The indicator \SmeasCol~indicates a collision when another vehicle overlaps with a static safety boundary of \SI{0.5}{\meter} around the ego vehicle. We introduce a static safety boundary instead of using the exact geometric collision check to account for potential inaccuracy due to sampling with \gls{mcts} approaches.

\subsection{Planning Algorithms}
We compare the \textbf{RC-RSBG} and the variant \textbf{RC-RSBGFullInfo} which employs the true simulated behavior policies \p{\otheridx} for prediction to several baseline algorithms.  We focus on interactive planners suitable for dense traffic. RC-RSBG and RC-RSBGFullInfo use a simplistic reward function $r(\cdot){=}1.0\cdot\textsc{goal reached}(\cdot)$. To model risk-awareness with the single-objective baselines, we define the reward function $r(\cdot){=}0.1\cdot\textsc{goal reached}(\cdot) - 0.1\frac{\SmeasEnv(\cdot)\cdot\tau_\text{predict}}{\SRisklevel\cdot\TPlan}   -1.0\SmeasCol(\cdot)$. It is designed such that it fully erases the goal reward when the predicted envelope violation duration $\sum_{\forall t^\prime}\SmeasEnv(\cdot)\tau_\text{predict}$, exceeds the allowed duration $\SRisklevel\cdot\TPlan$ with \TPlan~being the maximum planning horizon. 
Using this reward function, we define the following baselines
\begin{itemize}
	\item \textbf{RSBG} uses a \gls{mamcts} with hypothesis-based prediction and reward-based worst-case action selection for other agents, 
	\item \textbf{MDP} does not incorporate belief information over hypotheses. Instead it predicts other participants by sampling behavior states from the full hypothesized behavior space \BST{}{1D, \text{Head.}}{},
	\item \textbf{Cooperative} does not use the behavior space model for prediction. In its \gls{mamcts}, other agents select actions based on a global cost function which weights subjective and others' rewards based on a cooperation factor $c=0.1$ after evaluating $r(\cdot)$ for each agent. 
\end{itemize}
All baselines use an \gls{UCB} action selection strategy for the ego agent, the cooperative approach additionally for the other agents. All planners use an equal ego action space consisting of the macro actions lane changing, lane keeping at constant accelerations, $\dot{v_\egoidx}{=} \{-5, -1, -2, , 1, 2\} [\text{m}/\text{s}^2$] for freeway and $\dot{v_\egoidx}{=} \{-5, -1, 0, 1, 2\} [\text{m}/\text{s}^2$] for left turn, and gap keeping based on the IDM for freeway. The cooperative approach employs this action space additionally for the other agents. All planners use $\tau_a{=}\SI{0.2}{\second}$, a max search depth $d_\text{max}{=}10$ yielding $\TPlan{=}\SI{11}{\second}$. We obtain $\numhyp{=}16$ hypothesis $\p{\hbst{\hypidx}{}},\, \hypidx\in\{1,\ldots,\numhyp\}$ by equally partitioning the hypothesized behavior space $\BST{}{1D, \text{Head.}}{}$.  In this work, the focus is on evaluating the proposed risk formulation and decision model, the \gls{rcrsbg}, and not on working towards real-time capability with \gls{mamcts}. All planners use an equal number of 20000 iterations to minimize influence of sampling inaccuracies. 
\subsection{Results}
First, we \emph{qualitatively} analyze the \textbf{RC-RSBG} planner in Fig.~\ref{fig:qualitative_results} for two time steps in each scenario. The ego vehicle is shown in red. The posterior beliefs $\text{Pr}(\hbst{\hypidx}{}|\oh{t}, \otheridx)$ given for the four nearest other vehicles in the respective color qualitatively reflect the desired distance to the respective leading agent. The posterior is uniformly distributed for vehicles without a leading vehicle.  We see that the planned stochastic policy \p{\egoidx} correctly balances envelope and collision action-risk estimates, $\rho_\text{env}(\langle\oh{t}\rangle, \a{}{\egoidx}{})$ and $\rho_\text{col}(\langle\oh{t}\rangle, \a{}{\egoidx}{})$ such that the expected planned envelope risk $\rho_\text{env}^\text{exp.}$ (dashed red) fulfills the constraint $\SRisklevel=0.2$ (blue) while the expected planned collision risk $\rho_\text{col}^\text{exp.}$ (dotted red) is close to zero. Given these constraints, the planned policy prefers actions with higher expected action-return values $Q_R(\langle\oh{}{}\rangle, \a{}{\egoidx}{})$. Our qualitative analysis reveals that the \textbf{RC-RSBG} planner correctly implements the risk-constrained optimality criteria from Sec.~\ref{sec:rcrsbg} using a stochastic policy.

Next, we \emph{quantitatively} analyze the percentage of trials over 100 scenarios where the ego vehicle reaches the goal $P_\text{suc}$, collides $P_\text{col}$ or exceeds the maximum allowed simulation time, $P_\text{max}$ to solve a scenario ($t\leq 6.0\,[\text{s}]$). We evaluate over increasing envelope risk constraint $\SRisklevel$. For successful trials, we calculate the average time to reach the goal $\overline{T}_\text{suc}$. Results are given in Fig.~\ref{fig:barplot_results}. The \textbf{RC-RSBG} planner outperforms the baselines with increasing $\SRisklevel$ regarding $P_\text{suc}$ and $\overline{T}_\text{suc}$. The \textbf{MDP} and \textbf{Cooperative} planner do rarely succeed ($P_\text{suc} \ll 25\%$) emphasizing the advantage of belief-based prediction in denser traffic. With higher $\SRisklevel$ the \textbf{RC-RSBG} planner increasingly relies on the accuracy of the prediction model, and lesser on the safety provided by the envelope restriction. In the case of prediction model inaccuracies this provokes collisions for $\SRisklevel{\geq} 0.6$. This tendency is not observed with baseline planners which show collisions also for lower $\SRisklevel$. The \textbf{RC-RSBG-FullInfo} planner having access to the true behavior of other vehicles does not suffer from model inaccuracies, and, thus does not provoke any collision. These findings indicate the usefulness of multi-objective optimality to integrate risk constraints, and support our problem formulation considering risk to balance prediction model inaccuracies and safety, as discussed in Sec.~\ref{sec:problem_formulation}.    

To further analyze how $\SRisklevel$ balances safety and efficiency, we introduce two metrics, the observed envelope violation risk $\SRisklevel^{*}$ and the expected scenario waiting time $t_\text{w}$. The observed envelope violation risk is the percentage of simulation time the envelope is violated, $\SRisklevel^{*}= \frac{\sum_{\forall \scen}\sum_{\ost{t}{}\in \scen} \SmeasEnv(\ost{t}{}) \cdot \tau_a}{\sum_{\forall \scen}L(\scen)\cdot \tau_a}$ with $\ost{t}{}\in \scen$ giving the simulated states and $L(\scen)$ the length of the scenario \scen. The expected waiting time, $\overline{t}_\text{w}=\sum_{k=0}^{\infty}(\overline{T}_\text{max}k+\overline{T}_\text{suc})\cdot(P_\text{suc}P_\text{max}^k)$ defines the expected time to solve a scenario. The calculation assumes that the ego vehicle encounters solvable scenarios with probability $P_\text{suc}$ and duration $\overline{T}_\text{suc}$ and unsolvable scenarios with probability $P_\text{max}$ with duration equal to the allowed simulation time $\overline{T}_\text{max}=\SI{6}{\second}$. Fig.~\ref{fig:risk_mean} shows both metrics over risk level \SRisklevel.
In scenario freeway enter, the \textbf{RC-RSBG} planners fully exploit the allowed risk ($\SRisklevel^{*}\approx\SRisklevel$) for $\SRisklevel<0.8$, indicating that our \emph{risk formulation and planning approach reflects the observed risk}. For $\SRisklevel\geq 0.8$, $\SRisklevel^{*}$ decreases. We assume that the scenario ending becomes less risky by allowing higher $\SRisklevel$ initially. The opposite case occurs in the left turn scenario where a low $\SRisklevel$ prevents entering the intersection impeding full exploitation of allowed $\SRisklevel$.  The baseline approaches do not show any interpretable correlation between $\SRisklevel^{*}$ and $\SRisklevel$.
The waiting time $t_\text{w}$ indicates efficiency while $P_\text{col}$ from Fig.~\ref{fig:barplot_results}  defines safety. Interestingly, our risk formulation suggests $\SRisklevel \leq 0.4$ in the freeway enter scenario to prevent collisions which resembles the safety envelope violation risk of humans during lane changing, $\SRisklevel_\text{human} \approx 10\% - 40\%$ \cite{esterle_formalizing_2020, pek_verifying_2017} and introduces a natural trade off between safety and efficiency for $\SRisklevel > 0.4$. With our risk formulation waiting times are within a realistic human-like range, $t_\text{w}\approx\SI{2}{\second} \ldots \SI{15}{\second}$ in dense traffic whereas the baseline approaches show mostly larger waiting times.

We conclude, that the presented risk formulation and interactive planner balances safety and efficiency according to human-related risk criteria in dense traffic. It outperforms interactive baseline planners regarding efficiency and interpretability of their employed safety objectives.

\section{Conclusion}

This work formalizes a novel safety objective for interactive planning which balances safety and efficiency in dense traffic scenarios with uncertainty about other traffic participants' behavior by accepting a specifiable risk of violating a safety envelope. We propose a decision-theoretic framework under this safety objective, the \glsfirst{rcrsbg}, and an accompanying interactive planner based on a variant of Multi-Agent Monte Carlo Tree Search. In two types of traffic scenarios, we demonstrate that the \gls{rcrsbg} planner outperforms baseline planners and provides an interpretable and tunable safety objective.

This work reveals that a combination of uncertainty- and prediction based interactive planners with safety envelope restrictions is a promising direction for future research. We will further invest into the real-time capability of the planner, improve safety envelope definitions, and analyze in detail the correspondence between human risk concepts and the proposed risk level formalism.

\section{Acknowledgement}
This research was funded by the Bavarian Ministry of Economic Affairs, Regional Development and Energy, project Dependable AI.

\appendices


\AtNextBibliography{\footnotesize}
\printbibliography

\end{document}